%% file: survey.tex
\pdfoutput=1
\documentclass[journal,twoside]{IEEEtran}
\ifCLASSINFOpdf
\else
\fi

\usepackage{graphicx}
\usepackage{amsmath}
\usepackage{amssymb}
\usepackage[colorlinks]{hyperref}
\usepackage{verbatim}

\input{defs}

\usepackage{tabularx} 
\usepackage{booktabs}   
\usepackage{mathrsfs} 
\usepackage{multirow} 
\usepackage{amsfonts} 
\usepackage{wrapfig}  
\usepackage{cite}

\begin{document}

\title{Domain Adaptation for Medical Image Analysis: A Survey}

\author{Hao~Guan,
        and Mingxia~Liu,~\IEEEmembership{Senior Member,~IEEE}
\thanks{H. Guan and M. Liu are with the Department of Radiology and Biomedical Research Imaging Center, University of North Carolina at Chapel Hill, Chapel Hill, NC 27599, USA.}


\thanks{Corresponding author: M. Liu (mxliu@med.unc.edu).}
}


\if false
\markboth{IEEE TRANSACTIONS ON BIOMEDICAL ENGINEERING,~Vol.~XX, No.~XX, September~2020}%
{Hao \MakeLowercase{\textit{et al.}}: Domain Adaptation for Medical Image Analysis: A Survey}
\fi

\maketitle

\begin{abstract}
Machine learning techniques used in computer-aided medical image analysis usually suffer from the domain shift problem caused by different distributions between source/reference data and target data. 
As a promising solution, domain adaptation has attracted considerable attention in recent years. 
The aim of this paper is to survey the recent advances of domain adaptation methods in medical image analysis. 
We first present the motivation of introducing domain adaptation techniques to tackle domain heterogeneity issues for medical image analysis. 
Then we provide a review of recent domain adaptation models in various medical image analysis tasks. 
We categorize the existing methods into \emph{shallow} and \emph{deep} models, and each of them is further divided into \emph{supervised}, \emph{semi-supervised} and \emph{unsupervised} methods. 
We also provide a brief summary of the benchmark medical image datasets that support current domain adaptation research. 
This survey will enable researchers to gain a better understanding of the current status, challenges and future directions of this energetic research field.

\end{abstract}


%
\IEEEpeerreviewmaketitle

\section{Introduction}

\IEEEPARstart{M}{achine} learning has been widely used in medical image analysis, and typically assume that the training dataset (source/reference domain) and test dataset (target domain) share the same data distribution~\cite{learning}. 
However, this assumption is too strong and may not hold in real-world practice. 
Previous studies have revealed that the test error generally increases in proportion to the distribution difference between training and test datasets~\cite{test_error_1,test_error_2}. 
This is referred to as the well-known ``domain shift" problem~\cite{shift_1}. 
Even in the deep learning era, deep Convolutional Neural Networks (CNNs) trained on large-scale image datasets may still suffer from domain shift~\cite{decaf}. 
Thus, how to handle domain shift is a crucial issue to effectively apply machine learning methods to medical image analysis.

Unlike natural image analysis with large-scale labeled datasets such as ImageNet~\cite{ImageNet,ImageNet2}, in medical image analysis, a major challenge in constructing reliable and robust  machine learning models is the \emph{lack of labeled data}. 
Labeling medical images is generally expensive, time-consuming and tedious, requiring labor-intensive participation of physicians, radiologists and other experts. 
An intuitive solution is to reuse pre-trained models for some related domains~\cite{U-Net}. 
However, the domain shift problem is still widespread among different medical image datasets due to different scanners, scanning parameters, and subject cohorts, \etc. 
As a promising solution to tackle the domain shift/heterogeneity among medical image datasets, domain adaptation has attracted increasing attention in the medical image analysis community, aiming to minimize  distribution differences among different but related domains
Many researchers have engaged in leveraging domain adaptation methods to solve various tasks in medical image analysis. 

There have been a number of surveys on domain adaptation~\cite{DA_1,DA_2,DA_3,DA_4,DA_5,MDA_1,MDA_2} and transfer learning~\cite{TL_1,TL_2,TL_3,TL_4,TL_5,TL_6} with natural images. 
However, there are only very limited reviews related to domain adaptation and their applications in medical image analysis which is a broad and important research area. 
Cheplygina~\etal\cite{TL_MIA_1} provide a broad survey covering semi-supervised, multi-instance, and transfer learning for medical image analysis. 
Due to the wide scope, they only review general transfer learning methods in medical imaging applications without focusing on domain adaptation. 
Morid~\etal\cite{TL_MIA_2} review transfer learning methods based on pre-training on ImageNet~\cite{ImageNet} that can benefit from knowledge from large-scale labeled natural images. 
However, many recent research works tend to design 3D CNN-based domain adaptation models for medical image analysis which are not included in their survey.

In this paper, we review and discuss recent advances and challenges of domain adaptation for medical image analysis. 
We systematically summarize the existing methods according to their characteristics.
Specifically, we categorize different methods into two groups: 1) shallow models, and 2) deep models. Also, each group is further divided into three categories, \ie, supervised, semi-supervised and unsupervised methods. 
We also provide a summary of benchmark medical imaging datasets used in different domain adaptation research works.

The rest of this paper is organized as follows. We firstly introduce some background knowledge in Section~\ref{Background}. In Section~\ref{Shallow} and Section~\ref{Deep}, we review the recent advances of domain adaptation methods in medical image analysis. 
We then present benchmark medical image datasets in Section~\ref{Datasets}. 
Challenges and future research directions are discussed in Section~\ref{Discussions}. Finally, the conclusion is given in Section~\ref{Conclusion}.

\section{Background}    \label{Background}
\begin{figure}[!tbp]
\setlength{\belowcaptionskip}{-2pt}
\setlength{\abovecaptionskip}{-2pt}
\setlength{\abovedisplayskip}{-2pt}
\setlength{\belowdisplayskip}{-2pt}
\center
 \includegraphics[width= 1\linewidth]{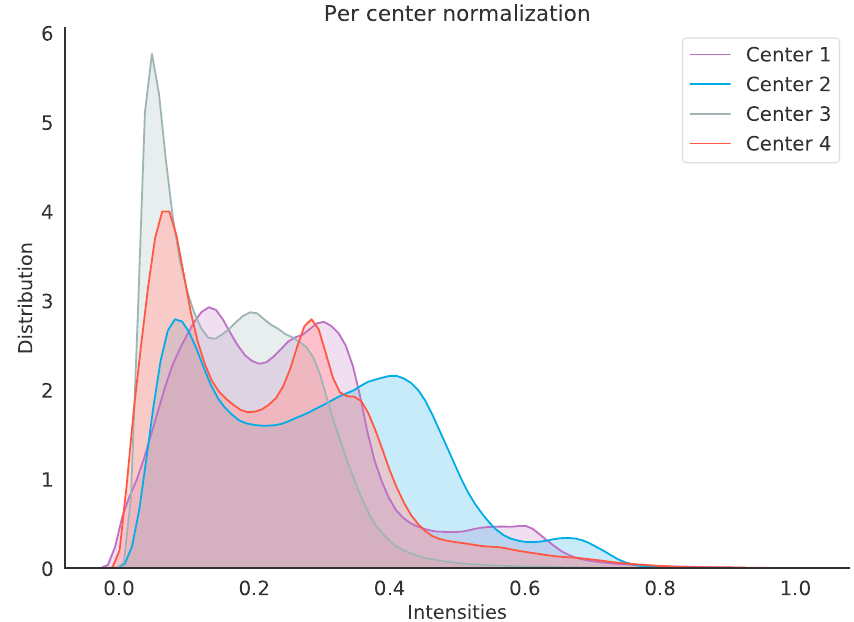}
 \caption{Intensity distribution of MRI axial-slice pixels from four different datasets (\ie, UCL, Montreal, Zurich, and Vanderbilt) that collected for gray matter segmentation. Intensity is normalized between 0 and 1 for each site. Image courtesy to C. Perone~\cite{Self_Ensemble_1}.}
 \label{fig:shift_1}
\end{figure}

\subsection{Domain Shift in Medical Image Analysis}
\begin{figure}[!tbp]
\setlength{\belowcaptionskip}{-2pt}
\setlength{\abovecaptionskip}{-2pt}
\setlength{\abovedisplayskip}{-2pt}
\setlength{\belowdisplayskip}{-2pt}
\center
 \includegraphics[width= 1\linewidth]{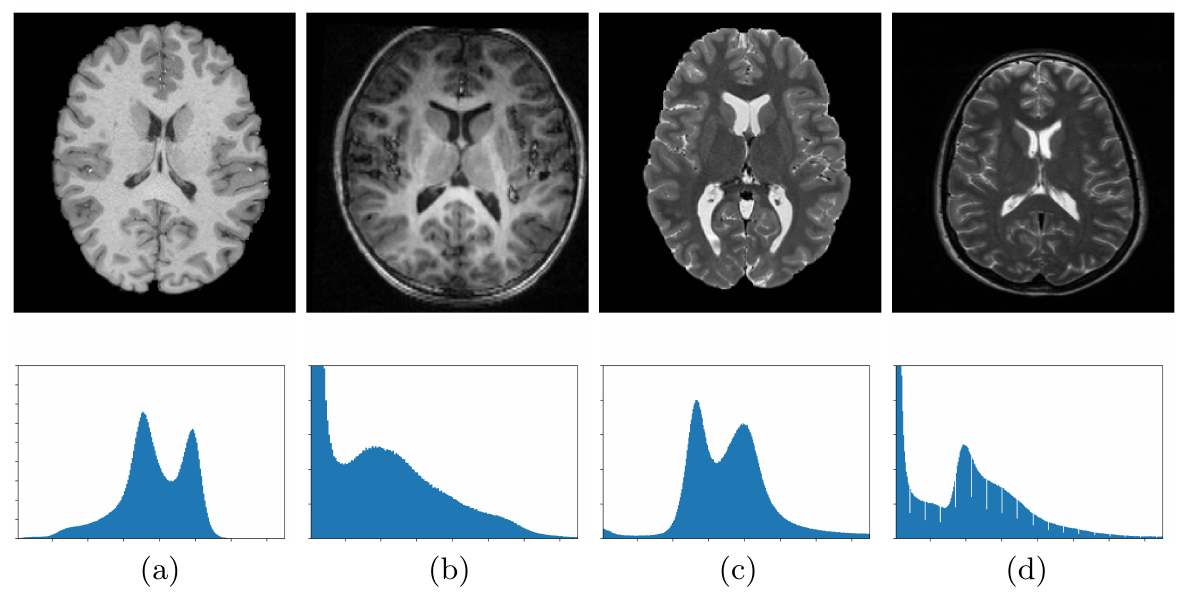}
 \caption{Image slices (top) and corresponding intensity distribution (bottom) of normalized T1-weighted (a, b) and T2-weighted (c, d) MRIs from different scanners. Image courtesy to N. Karani~\cite{Multi_DA_BN}.}
 \label{fig:shift_2}
\end{figure}

For a machine learning model, domain shift~\cite{shift_1,shift_2,shift_3,shift_4} refers to the change of data distribution between its training dataset (source/reference domain) and test dataset (target domain). 
The domain shift problem is very common in practical applications of various machine learning algorithms and may cause significant performance degradation. 
Especially for multi-center studies, domain shift is widespread among different imaging centers/sites that may use different scanners, scanning protocols, and subject populations, \etc. 
Fig.~\ref{fig:shift_1} illustrates the problem of inter-center domain shift in terms of the intensity distribution of structural magnetic resonance imaging (MRI) from four independent sites (\ie, UCL, Montreal, Zurich, Vanderbilt) in the Gray Matter segmentation challenge~\cite{Self_Ensemble_1,Fig_1}. 
Fig.~\ref{fig:shift_2} illustrates the image-level distribution heterogeneity caused by different scanners~\cite{Multi_DA_BN}. From Figs.~\ref{fig:shift_1}-\ref{fig:shift_2}, one can observe that there are clear distribution shifts in these medical imaging data. 
However, many conventional machine learning methods ignore this problem, which would lead to performance degradation~\cite{shift_3,shift_5}. 
Recently, domain adaptation has attracted increasing interests and attention of researchers, and become an important research topic in machine learning based medical image analysis~\cite{DA_Evaluation,Transfusion,DANN_Seg,TL_MIA_1}.

\subsection{Domain Adaptation and Transfer Learning}
This survey focuses on domain adaptation for medical image analysis. 
Since domain adaptation can be regarded as a special type of transfer learning, we first review their definitions to provide a clear understanding of their differences. 
In a typical transfer learning setting, there are two concepts: ``domain'' and ``task''~\cite{TL_1,TL_2,TL_3}. 
A domain relates to the feature space of a specific dataset and the marginal probability distribution of features. 
A task relates to the label space of a dataset and an objective predictive function. 
The goal of transfer learning is to transfer the knowledge learned from the task $T_a$ on domain $A$ to the task $T_b$ on domain $B$. Note that either the domain and the task may change during the transfer learning process.

Domain adaptation, the focus of this survey, is a particular and popular type of transfer learning~\cite{DA_1,DA_2,DA_3,DA_4,DA_5}. 
For domain adaptation, it is assumed that the domain feature spaces and tasks remain the same while the marginal distributions are different between the source and target domains (datasets). 
It can be mathematically depicted as follows.
Let $\mathcal{X} \times \mathcal{Y}$ represent the joint feature space and the corresponding label space, respectively. 
A source domain {S} and a target domain {T} are defined on $\mathcal{X} \times \mathcal{Y}$, with different distributions $P_s$ and $P_t$, respectively. 
Suppose we have $n_s$ labeled samples (subjects) in the source domain, \ie, $\mathcal{D}_S = \{(\mathbf{x}^S_i, y^S_i)\}^{n_s}_{i=1}$, and also have $n_t$ samples (with or without labels) in the target domain, \ie, $\mathcal{D}_T = \{(\mathbf{x}^T_j)\}^{n_t}_{j=1}$. 
Then the goal of domain adaptation (DA) is to transfer knowledge learned from {S} to {T} to perform a specific task on {T}, and this task is shared by {S} and T.

\begin{figure*}[!t]
\setlength{\belowcaptionskip}{-2pt}
\setlength{\abovecaptionskip}{-2pt}
\setlength{\abovedisplayskip}{-2pt}
\setlength{\belowdisplayskip}{-2pt}
\begin{center}
\includegraphics[width=0.96\linewidth]{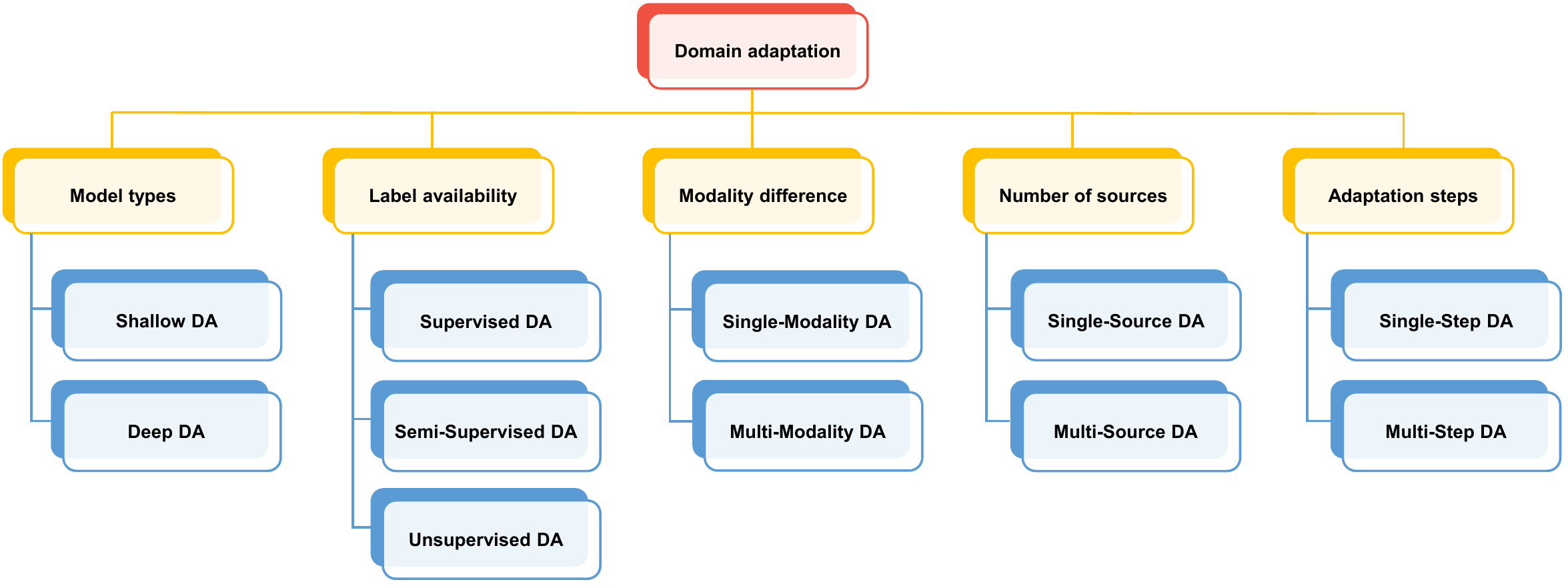}
\end{center}
   \caption{An overview of different categories of domain adaptation methods for medical image analysis.}
\label{fig:main}
\end{figure*}

\subsection{Different Settings of Domain Adaptation}
In DA, the source domain and target domain share the same learning tasks. 
In practice, DA can be categorized into different groups according to different scenarios, constrains, and algorithms. 
In Fig.~\ref{fig:main}, we summarize different categories of DA methods for medical image analysis based on six problem settings, \ie, model type, label availability, modality difference, number of sources, and adaptation steps. 
\if false
It should be noted that each group of methods are not mutually exclusive to each other.  
For example, there are also supervised, semi-supervised and unsupervised DA methods in shallow and deep DA approaches. 
In this survey, we employ the model type, \ie, shallow learning or deep shallow learning, as the main criterion for categorization to review the existing DA methods for medical image analysis. 
\fi

\begin{itemize}
\item \emph{Model Type}: \textbf{Shallow DA \& Deep DA}. In terms of whether the learning model is shallow or deep, the DA methods can be divided into shallow DA and deep DA~\cite{DA_5,DA_2,DA_1}. 
Shallow DA methods usually rely on human-engineered imaging features and conventional machine learning models. 
In contrast, deep DA methods (especially those with CNN architectures) generally integrate feature learning and model training into end-to-end learning models, where the data adaptation is performed in a task-oriented manner. 

\item \emph{Label Availability}: \textbf{Supervised DA \& Semi-Supervised DA \& Unsupervised DA}. In terms of label availability, existing DA methods can be divided into supervised DA, semi-supervised DA, and unsupervised DA~\cite{DA_5,DA_3}.
In supervised DA, a small number of labeled data in the target domain are available for model training.
In semi-supervised DA, a small number of labeled data as well as redundant unlabeled data in the target domain are available in the training process.
In unsupervised DA, only unlabeled target data are available for training the adaptation model. 
Since label scarcity is widespread among medical images, unsupervised DA has attracted increasing attention recently.
In fact, there is an extreme case that even unlabeled target data are not accessible during the training time. 
The learning model is only allowed to be trained on several related source domains to gain enough generalization ability for the target domain. 
This problem is referred to as \textbf{domain generalization}, while a few related research has been conducted for medical image analysis~\cite{DA_G} currently.

\item \emph{Modality Difference}: \textbf{Single-Modality DA \& Cross-Modality DA}. In terms of modality difference, the existing DA methods can be divided into single-modality DA and cross-modality DA~\cite{DA_1,Disentangle,SIFA}. 
In single modality DA, the source and target domains share the same data modality. For example, the source domain consists of MRIs collected in vendor A, and the target domain contains MRIs acquired from vendor B~\cite{Unet-GAN}.
In cross-modality DA, the modality of source and target domains are different with various scanning technologies. For example, the source domain consists of MR images, whereas the target domain contains CT images~\cite{SIFA}.

\item \emph{Number of Sources}: \textbf{Single-Source DA \& Multi-Source DA}. In terms of the number of source domains, the existing DA methods can be divided into single-source DA and multi-source DA approaches~\cite{DA_4,MDA_1,MDA_2}. 
In single-source DA, the model is trained on a single source domain, whereas multi-source DA uses training samples from multiple source domains. 
Since there is also data heterogeneity among different source domains, multi-source DA is quite challenging. 
Most of the existing DA methods for medical image analysis fall into the single-source DA category.

\item \emph{Adaptation Step}: \textbf{One-Step DA \& Multi-Step DA}. In terms of adaptation steps, the existing DA methods can be divided into one-step DA and multi-step DA~\cite{DA_2,Multi_Step_1,Multi_Step_2}. 
In one-step DA, adaption between the source and target domain is accomplished in one step due to a relatively close relationship between them. 
In multi-step DA (also called distant/transitive DA), there are often intermediate domains introduced to bridge the relatively large distribution gap between the source and target domains. 
\end{itemize}

It should be noted that each group of methods are not mutually exclusive to each other. 
For example, there are also supervised, semi-supervised and unsupervised DA methods in shallow and deep DA approaches. 
In this survey, we employ the model type as the main criterion for DA method categorization, and introduce existing shallow DA and deep DA methods for medical image analysis. 

\section{Shallow Domain Adaptation Methods}   \label{Shallow}
\subsection{Overview}
In this section, we review shallow domain adaptation methods based on human-engineered features and conventional machine learning models for medical image analysis. 
We first introduce two commonly-used strategies in shallow DA methods: 1) instance weighting, and 2) feature transformation. 

\begin{figure}[!tbp]
\setlength{\belowcaptionskip}{-2pt}
\setlength{\abovecaptionskip}{-2pt}
\setlength{\abovedisplayskip}{-2pt}
\setlength{\belowdisplayskip}{-2pt}
\center
 \includegraphics[width= 1\linewidth]{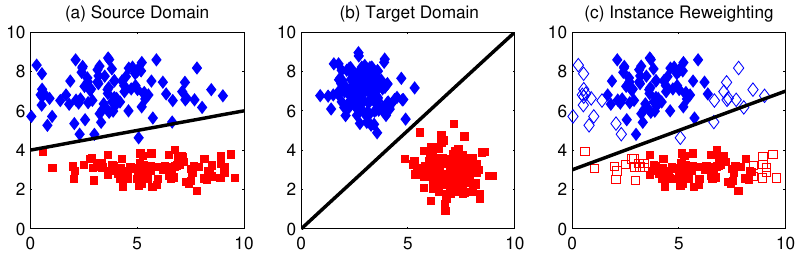}
 \caption{Illustration of instance weighting which can alleviate domain shift. (a) Source domain after feature matching (\ie, discovering a shared feature representation by jointly reducing the distribution difference and preserving the important properties of input data). (b) Target domain after feature matching. (c) Source domain after joint feature matching and instance weighting, with unfilled markers indicating irrelevant source instances that have smaller weights. Image courtesy to Long~\etal~\cite{Joint_Long}.}
 \label{fig:instance_weighting}
\end{figure}

Instance weighting is one of the most popular strategies adopted by shallow DA methods for medical image analysis~\cite{Reweighting_Wachinger,re-weighting,Opbroek_1,Lung_Weighting,Weighting_Ultrasound}. 
In this strategy, samples/instances in the source domain are assigned with different weights according to their relevance with target samples/instances. 
Generally, source instances that are more relevant to the target instances will be assigned larger weights. 
After instance weighting, a learning model (\eg, classifier or regressor) is trained on the re-weighted source samples, thus reducing domain shift between the source and target domains.
Fig.~\ref{fig:instance_weighting} illustrates the effect of instance weighting strategy on the source and target domains. 
From Fig.~\ref{fig:instance_weighting}(a)-(b), we can see that the domain difference between the source and target domain is large. After instance weighting, the domain difference is reduced based on those re-weighted source instances, as shown in Fig.~\ref{fig:instance_weighting}(b)-(c). 

\begin{figure}[!tbp]
\setlength{\belowcaptionskip}{-2pt}
\setlength{\abovecaptionskip}{-2pt}
\setlength{\abovedisplayskip}{-2pt}
\setlength{\belowdisplayskip}{-2pt}
\center
 \includegraphics[width= 1\linewidth]{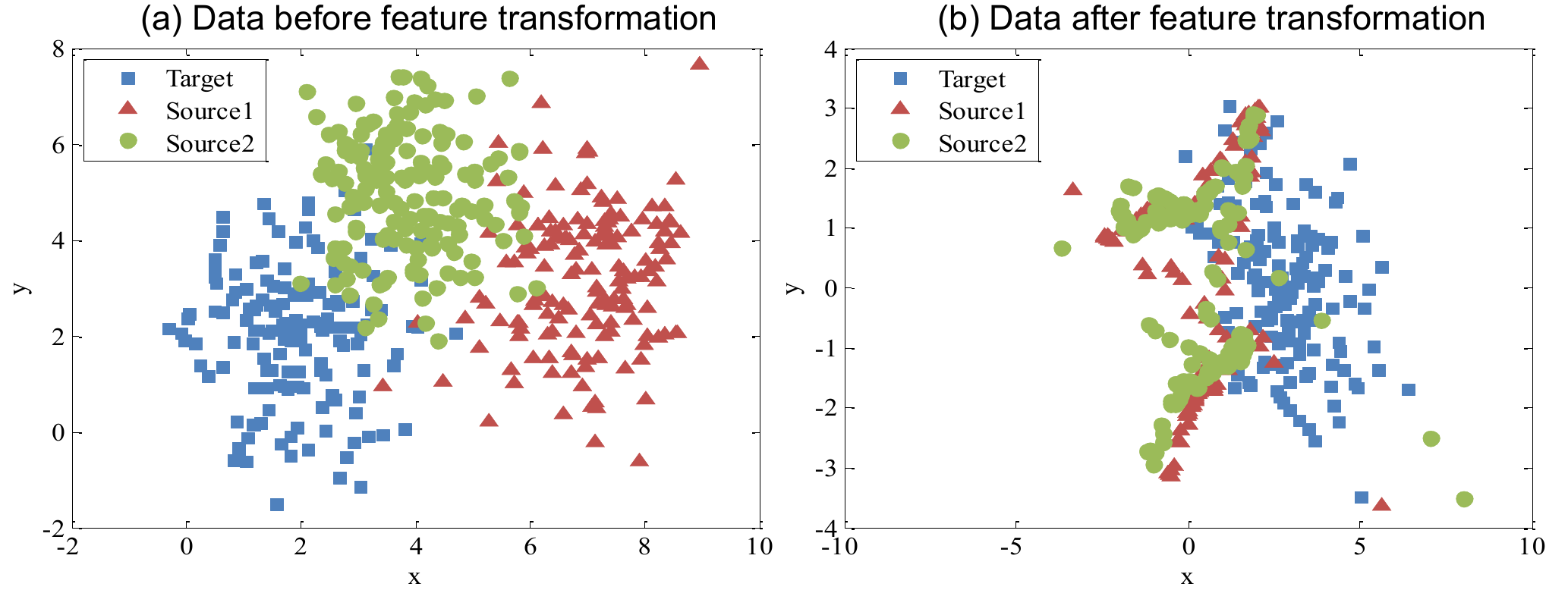}
 \caption{Illustration of feature transformation to reduce domain shift. Each color denotes a specific domain. (a) Data distributions of two source and one target domains before feature transformation, and (b) data distributions after feature transformation. Image courtesy to Wang~\etal~\cite{ASD_MDA_2}.}
 \label{fig:featureTransformation}
\end{figure}

Feature transformation strategy focuses on transforming source and target samples from their original feature spaces to a new shared feature representation space~\cite{Latent_Becker,ASD_MDA_1,ASD_MDA_2}.
As shown in Fig.~\ref{fig:featureTransformation}, the goal of feature transformation for DA is to construct a common/shared feature space for the source and target domains to reduce their distribution gap, based on various techniques such as low-rank representation~\cite{ASD_MDA_2}. 
Then, a learning model can be trained on the new feature space, which is less affected by the domain shift in the original feature space between the two domains.

\subsection{Supervised Shallow DA}
Wachinger~\etal\cite{Reweighting_Wachinger} propose an instance weighting-based DA method for Alzheimer's disease (AD) classification. 
Specifically, labeled target samples are used to estimate the target distribution.
Then source samples are re-weighted by calculating the probability of source samples in the target domain. 
A multi-class logistic regression classifier is finally trained on the re-weighted samples and labeled target data (with volume, thickness, and anatomical shape features), and applied to the target domain.
Experiments on the ADNI~\cite{ADNI}, AIBL~\cite{AIBL} and CADDementia challenge datasets~\cite{CADDementia} show that the DA model can achieve better results than methods that only use data from either the source or target domain.
Goetz~\etal\cite{re-weighting} propose an instance weighting-based DA method for brain tumor segmentation. 
To compute the weights of source samples, a domain classifier (logistic regression) is trained with paired samples from source and target domains. 
The output of the domain classifier (in terms of probability) is then used to calculate the adjusted weights for source samples. A random forest classifier is trained to perform segmentation.
Experiments on the BraTS dataset~\cite{BRATS} demonstrate the effectiveness of this method.
Van Opbroek~\etal\cite{Opbroek_1} propose an instance weighting-based DA method for brain MRI segmentation. 
In their approach, each training image in the source domain is assigned a weight that can minimize the difference between the weighted probability density function (PDF) of voxels in the source and target data. 
The re-weighted training data are then used to train a classifier for segmentation. 
Experiments on three segmentation tasks (\ie, brain-tissue segmentation, skull stripping, and white-matter-lesion segmentation) demonstrate its effectiveness.

Becker~\etal\cite{Latent_Becker} propose a feature transformation-based DA method for microscopy image analysis. Specifically, they propose to learn a nonlinear mapping that can project samples in both domains to a shared discriminative latent feature space. 
Then, they develop a boosting classifier trained on the data in the transformed common space.
In~\cite{DTSVM_1,DTSVM_2}, the authors propose a domain transfer support vector machine (DTSVM) for mild cognitive impairment (MCI) classification. 
Based on the close relationship of Alzheimer's disease (AD) and MCI, DTSVM is trained on MCI subjects in the target domain and fine-tuned with an auxiliary AD dataset to enhance its generalization ability. 
They validate their method on the ADNI database and achieve good performance in MCI conversion prediction.

\subsection{Semi-Supervised Shallow DA}
Conjeti~\etal\cite{Random_Forest} propose a two-step DA framework for ultrasound image classification. 
In the first step, they use principle component analysis (PCA) to transform the source and target domains to a common latent space, through which the global domain difference is minimized. 
In the second stage, a random forest classifier is trained on the transformed source domain, and then fine-tuned with a few labeled target data to further reduce the domain shift. 

\subsection{Unsupervised Shallow DA}
Cheplygina~\etal\cite{Lung_Weighting} employ the instance weighting strategy for lung disease classification. 
Specifically, Gaussian texture features are first extracted from all CT images, based on which a weighted logistic classifier is trained. Source samples that are similar to the target data are assigned high weights to reduce domain shift. 
Experiments on four chest CT image datasets show that this method can significantly improve the classification performance.
Heimann~\etal\cite{Weighting_Ultrasound} use instance weighting-based DA strategy for ultrasound transducer localization in X-ray images. 
The instance weights are computed through a domain classifier (\ie, logistic regression). 
Based on Haar-like features of X-ray images, a cascade of tree-based classifiers is trained for ultrasound transducer localization. 
Li~\etal\cite{AD_SVD} propose a subspace alignment DA method for AD classification using functional MRI (fMRI) data. 
They first conduct feature extraction and selection for source and target samples, followed by a modified subspace alignment strategy to align samples from both domains into a shared subspace. 
Finally, the aligned samples in the shared subspace are used as an integrated dataset to train a discriminant analysis classifier.
This method is validated on both ADNI and a private fMRI datasets.
Kamphenkel~\etal\cite{DKI} propose an unsupervised DA method for breast cancer classification based on diffusion-weighted MR images (DWI). 
They use the Diffusion Kurtosis Imaging (DKI) algorithm to transform the target data to the source domain without any label information of target data.

\subsection{Multi-Source Shallow DA}
Wang~\etal\cite{ASD_MDA_1} propose a multi-source DA framework for Autism Spectrum Disorder (ASD) classification with resting-state functional MRI (rs-fMRI) data. 
Each subject/sample is represented by both functional connectivity (FC) features in gray matter regions and functional correlation tensor features extracted from white matter regions.
All subjects in each of the multiple source domains are transformed to the target domain via low-rank regularization and graph embedding. 
Then, a classifier is trained on the source samples (with transformed features). 
This method is validated on the ABIDE dataset~\cite{ABIDE}. 
Wang~\etal\cite{ASD_MDA_2} propose a low-rank representation based multi-source domain adaptation framework for ASD classification.
They assume that multiple domains share an intrinsic latent data structure, and map multiple source data and target data into a common latent space via a low-rank representation to reduce the domain shift.
Experimental results on both synthetic data and the ABIDE dataset suggest the good performance of their method. 
Cheng~\etal\cite{Multiple_Domain_MCI} propose a Multi-Domain Transfer Learning (MDTL) framework for the early diagnosis of AD and MCI. 
They first select a subset of discriminative features (\ie, gray matter tissue volume) by using the training data from multiple auxiliary domains and target domain. 
Then, they construct a multi-domain transfer classifier based on the ensemble of the classifiers learned from auxiliary domains and an adaptive SVM~\cite{ASVM} learned from the labeled target samples.

\section{Deep Domain Adaptation Methods}  
\label{Deep}
Deep learning~\cite{DL,DL2} has greatly pushed forward the development of artificial intelligence and machine learning. 
As one of the most popular deep learning models, CNN has demonstrated its superiority over conventional human-engineered imaging features.
Trained with large-scale labeled data in a full supervision manner, CNN has made breakthroughs in computer vision and medical image analysis~\cite{DL_Medical,DL_Medical2,DL_Medical3}. 
It has been revealed that CNN is able to learn generic low-level features (\eg, textures and edges) that can be transferable to different image analysis tasks~\cite{CNN-transfer,CNN-visual}.
A milestone is the AlexNet~\cite{Alexnet} which is trained with millions of natural images from ImageNet~\cite{ImageNet}. 
Following AlexNet, several representative CNN models have been proposed, including VGG~\cite{VGG}, Inception~\cite{Inception}, ResNet~\cite{ResNet}, and DenseNet~\cite{DenseNet}. 
These four types of CNNs have been widely used in various domain adaptation models for medical image analysis~\cite{TL_MIA_2}.

\subsection{Supervised Deep DA}
With deep features (\eg, extracted by CNNs), several studies focus on using shallow DA models for medical image analysis. 
Kumar~\etal\cite{ResNet_TCA} use ResNet as the feature extractor of mammographic images. Based on the CNN features, they evaluate three shallow domain adaptation methods, \ie, Transfer Component Analysis (TCA)~\cite{TCA}, Correlation Alignment (CORAL)~\cite{CORAL}, and Balanced Distribution Adaptation (BDA)~\cite{BDA} for breast cancer classification, and provide some empirical results on the DDSM~\cite{DDSM} and InBreast~\cite{InBreast} datasets. 
Huang~\etal\cite{Huang} propose to use LeNet-5 to extract features of histological images from different domains for epithelium-stroma classification, and then project them into a subspace (via PCA) to align them for adaptation. Experiments on the NKI, VGH~\cite{NKI_VGH} and IHC~\cite{IHC} datasets verify its effectiveness.

Another research direction is to transfer models learned on the source domain onto the target domain with fine-tuning. 
Ghafoorian~\etal\cite{DA_Evaluation} evaluate the impact of fine-tuning strategy on brain lesion segmentation, based on CNN models pre-trained on brain MRI scans. 
Their experimental results reveal that using only a small number of target training examples for fine-tuning can improve the transferability of models. 
They further evaluate the influence of both the size of the target training set and different network architectures on the adaptation performance.
%
Based on similar findings, numerous methods have been proposed to leverage CNNs that are well pre-trained on ImageNet to tackle medical image analysis problems.
Samala~\etal\cite{Samala} propose to first pre-train an AlexNet-like network on ImageNet, and then fine-tune it with regions-of-interest (ROI) from 2,454 mass lesions for breast cancer classification.
Khan~\etal\cite{Khan} propose to pre-train a VGG network on ImageNet, and then use labeled MRI data to fine-tune it for Alzheimer's disease (AD) classification. 
They propose to use image entropy to select the most informative training samples in the target domain, and verify their method on the ADNI database~\cite{ADNI}.
Similarly, Swati~\etal\cite{VGG_MRI} pre-train a VGG network on  ImageNet and re-train the higher layers of the network with labeled MR images for brain tumor classification. 
Experiments are conducted on a publicly available CE-MRI dataset\footnote{https://figshare.com/articles/brain\_tumor\_dataset/1512427}. 
%
Abbas~\etal\cite{Class_Decompose} employ the ImageNet to pre-train CNNs for chest X-ray classification. To deal with the irregularities in the datasets, they introduce class decomposition into the network learning process, by partitioning each class within the image dataset into $k$ subsets and then assign new labels to the new set. 
Three different cohorts of chest X-ray images, histological images of human colorectal cancer, and digital mammograms are used for performance evaluation. 

The above-mentioned methods employ the one-step DA strategy, \ie, directly transferring the pre-trained model to the target domain. 
When facing the problem setting where target samples are too few to fine-tune a model, several studies propose to use intermediate domains to facilitate multi-step DA. 
Gu~\etal\cite{Two_Step_TL} develop a two-step adaptation method for skin cancer classification. First, they fine-tune a ResNet on a relatively large medical image dataset (for skin cancer). 
Then, the network is trained on the target domain which is a relatively small medical image dataset. 
Their experimental results on the MoleMap dataset\footnote{http://molemap.co.nz} and the HAM10000 dataset\footnote{https://www.kaggle.com/kmader/skin-cancer-mnist-ham10000} show that the two-step adaptation method achieves better results than the directly transferring methods.

Although adopting pre-trained models on a dataset (\eg, ImageNet) is a popular way for supervised DA for medical image analysis, the 2D CNN structure may not able to fully explore the rich information conveyed in 3D medical images. 
To this end, some researchers deliberately design task-specific 3D CNNs that are trained with medical images as the backbone to facilitate the subsequent data adaptation tasks. 
Hosseini-Asl~\etal\cite{Adaptive_CNN} design a 3D CNN for brain MR images classification. 
The network is pre-trained with MR images in the source domain. Then, its upper fully-connected layers are fine-tuned with samples in the target domain. Experiments on ADNI and CADDementia demonstrate its effectiveness.
Valverde~\etal\cite{FC_DA} propose a similar 3D CNN for brain MR images segmentation. 
Instead of using the whole brain MRI, their network takes 3D image patches as input, while only partial fully-connected layers are fine-tuned using the target data. This method is evaluated on the ISBI2015 dataset~\cite{ISBI2015} for MRI segmentation.
Kaur~\etal\cite{Different_Disease_TL} propose to first pre-train a 3D U-Net on a source domain that has relevant diseases with a large number of samples, and then use a few labeled target data to fine-tune the network. Experiments on the BraTS dataset~\cite{BRATS} show this strategy achieves a better performance than the network trained from scratch.
Similar strategy is also used in~\cite{Different_Data_TL} where a network is pre-trained using a large number of X-ray computed tomography (CT) and synthesized radial MRI datasets and then fine-tuned with only a few labeled target MRI scans.
Zhu~\etal\cite{BOWDA} propose a boundary-weighted domain adaptive neural network for prostate segmentation. 
A domain feature discriminator is co-trained with the segmentation networks in an adversarial learning manner to reduce domain shift.
Aiming to tackle the difficulty caused by unclear boundaries, they design a boundary-weighted loss and add it into the training process for DA and segmentation. 
The boundary contour needs to be extracted from the ground truth label to generate a boundary-weighted map as the supervision information to minimize the loss. 
Experiments on the PROMISE12 challenge dataset\footnote{https://promise12.grand-challenge.org/} and BWH dataset~\cite{BWH} demonstrate its effectiveness. 
%
Bermúdez-Chacón~\etal\cite{Two_Unet} design a two-stream U-Net for electron microscopy image segmentation. 
One stream uses source domain samples while the other uses target data. They utilize Maximum Mean Discrepancy (MMD) and correlation alignment as the domain regularization for DA. With a few labeled target data to fine-tune the model, their method achieves promising performance in comparison with several state-of-the-art ones on a private dataset.
Laiz~\etal\cite{Triplet} propose to use triplet loss for DA in endoscopy image classification. 
Each triplet consists of an anchor sample A from the source domain, a positive sample B from the target domain with the same label of A, and a negative sample C in the source domain. 
Through minimizing the triplet loss, their model can reduce the domain shift while keeping discrimination on different diseases.

\subsection{Semi-Supervised Deep DA}
Roels~\etal\cite{Y_Net} propose a semi-supervised DA method for electron microscopy image segmentation. 
They design a ``Y-Net'' with one feature encoder and two decoders. One decoder is used for segmentation, while a reconstruction decoder is designed to reconstruct images from both the source and target domains. The network is initially trained in an unsupervised manner. Then, the reconstruction decoder is discarded, and the whole network is fine-tuned with labeled target samples to make the model adapt to the target domain.
 
Madani~\etal\cite{Semi_GAN} propose a semi-supervised generative adversarial network (GAN) based DA framework for chest X-ray image classification. 
Different from conventional GAN, this model takes labeled source data, unlabeled target data and generated images as input. 
The discriminator performs three-category classification (\ie, normal, disease, or generated image). 
During training, unlabeled target data can be classified as any of those three classes, but can contribute to loss computation when they are classified as generated images. 
Through this way, both labeled and unlabeled data can be incorporated into a semi-supervised manner. 
Experiments on the NIH PLCO dataset~\cite{PLCO} and NIH Chest X-Ray dataset~\cite{NIH_Chest} demonstrate its effectiveness.

\subsection{Unsupervised Deep DA}
\begin{figure}[!tbp]
\setlength{\belowcaptionskip}{-2pt}
\setlength{\abovecaptionskip}{-2pt}
\setlength{\abovedisplayskip}{-2pt}
\setlength{\belowdisplayskip}{-2pt}
\center
 \includegraphics[width= 1\linewidth]{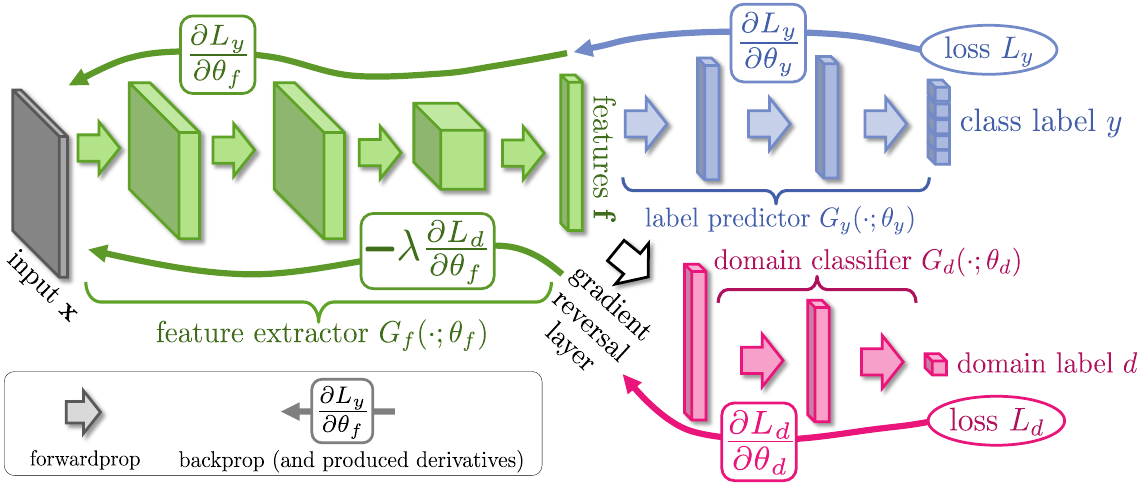}
 \caption{Illustration of Domain Adversarial Neural Network (DANN) framework. It is a classic and efficient DA model for domain-invariant feature learning through adversarial training. Image courtesy to Ganin~\etal~\cite{DANN}.}
 \label{fig:DANN}
\end{figure}

Unsupervised deep domain adaptation has attracted increasing attention~\cite{DA_3} in the field of medical image analysis, due to its advantage that does not require any labeled target data. We now introduce the existing unsupervised deep DA methods based on their specific strategies for knowledge transfer. 

\subsubsection{Feature Alignment}
This line of research aims to learn domain-invariant features across domains through specifically designed CNN models. Most of the models adopt a siamese architecture similar to the Domain Adversarial Neural Network (DANN) structure~\cite{DANN} as shown in Fig.~\ref{fig:DANN}. Kamnitsas~\etal\cite{DANN_Seg} propose a DANN-based multi-connected adversarial network for brain lesion segmentation. 
In their model, the domain discriminator is trained simultaneously with a segmentation network. In addition, the authors argue that only adapting the last layers of the segmentor is not ideal, thus the domain discriminator is connected at multiple layers of the network to make it less susceptible to image-quality variations between different domains.
Javanmardi~\etal\cite{DANN_Eye} propose a DANN-based model for eye vasculture segmentation. 
A U-Net and a domain discriminator are co-trained during the training process. Experiments on the DRIVE~\cite{DRIVE} and STARE~\cite{STARE} datasets demonstrate its effectiveness.
Yang~\etal\cite{ROI_DANN} employ the DANN for lung texture classification, with experiments performed on the SPIROMICS dataset\footnote{https://www.spiromics.org/spiromics/}.
Panfilov~\etal\cite{Knee_MRI} develop an adversarial learning-based model for knee tissue segmentation. A U-Net-based segmentor and a domain discriminator with adversarial learning are co-trained for DA. Experiments on three knee MRI datasets verify its effectiveness.
%
Zhang~\etal\cite{UCAN} propose an adversarial learning-based DA for AD/MCI classification on ADNI. A classifier and a domain discriminator are co-trained to enhance the model's transferability across different domains. 

Different from the above-mentioned studies, Dou~\etal\cite{Early_DA} develop a cross-modality DA framework for cardiac MR and CT image segmentation, by only adapting low-level layers (with higher layers fixed) to reduce domain shift during training.
They assume that the data shift between cross-modality domains mainly lies in low-level characteristics, and validate this assumption through experiments on the MM-WHS dataset~\cite{MM-WHS}.
%
Shen~\etal\cite{ADDA_Eye} employ adversarial learning for fundus image quality assessment. Due to the demand for image assessment, they fix high-level weights during the adversarial training to focus on low-level feature adaptation. 
%
Yan~\etal\cite{Canny_DA} propose an adversarial learning-based DA method for MR image segmentation. A domain discriminator is co-trained with the segmentor to learn domain-invariant features for the task of segmentation.
To enhance the model's attention to edges, Canny edge detector is introduced into the adversarial learning process. Experiments on images from three independent MR vendors (Philips, Siemens, and GE) demonstrate the effectiveness. 
%
Shen~\etal\cite{mammogram_Shen} propose an adversarial learning-based method for mammogram detection which is an essential step in breast cancer diagnosis. A segmentor based on fully convolutional network (FCN) and a domain discriminator are co-trained for domain adaptation.
Experiments on the public CBIS-DDSM~\cite{DDSM}, InBreast~\cite{InBreast} and a self-collected dataset demonstrate its effectiveness.
%
Yang~\etal\cite{RPN_local_DA} propose an adversarial learning-based DA method for lesion detection within the Faster RCNN framework~\cite{Faster_RCNN}. 
Besides global feature alignment, they also extract ROI-based region proposals to facilitate local feature alignment. Through combining these two level features, their method shows robustness in the experiments on two datasets from two independent medical centers.

Gao~\etal\cite{fMRI_CMD} propose an unsupervised method for classification of brain activity based on fMRI data from the Human Connectome Project (HCP) dataset~\cite{HCP}. 
They use the central moment discrepancy (CMD) which matches the higher-order central moments of the data distributions as the domain regularization loss to facilitate adaptation. 
They assume that the high-level features extracted by fully-connected layers have large domain shift, thus CMD is imposed on these layers to perform adaptation. 
%
Bateson~\etal\cite{Constrained_DA} propose an unsupervised constrained DA framework for disc MR image segmentation. 
They propose to use some useful prior knowledge that is invariant across domains as an inequality constraint, and impose such constraints on the 
predicted results of unlabeled target samples as the domain adaptation regularization. 
In practice, they employ the size of target segmentation regions as a constraint for network optimization, with experiments performed on the MICCAI 2018 IVDM3Seg Challenge\footnote{https://ivdm3seg.weebly.com/}. 
Mahapatra~\etal\cite{register_GAN} develop a deep DA approach for cross-modality image registration. 
To transfer knowledge across domains, they design a convolutional auto-encoder to learn latent feature representations of images, followed by a generator to synthesize the registered image. 
Trained on the latent feature space instead of on the original images, their model is less limited to any imaging modality. 
Experiments on chest X-ray, retinal and brain MR images show that their model (trained on one dataset) gives better registration performance for other datasets. 

\subsubsection{Image Alignment}
\begin{figure}[!tbp]
\setlength{\belowcaptionskip}{-2pt}
\setlength{\abovecaptionskip}{-2pt}
\setlength{\abovedisplayskip}{-2pt}
\setlength{\belowdisplayskip}{-2pt}
\center
 \includegraphics[width= 1\linewidth]{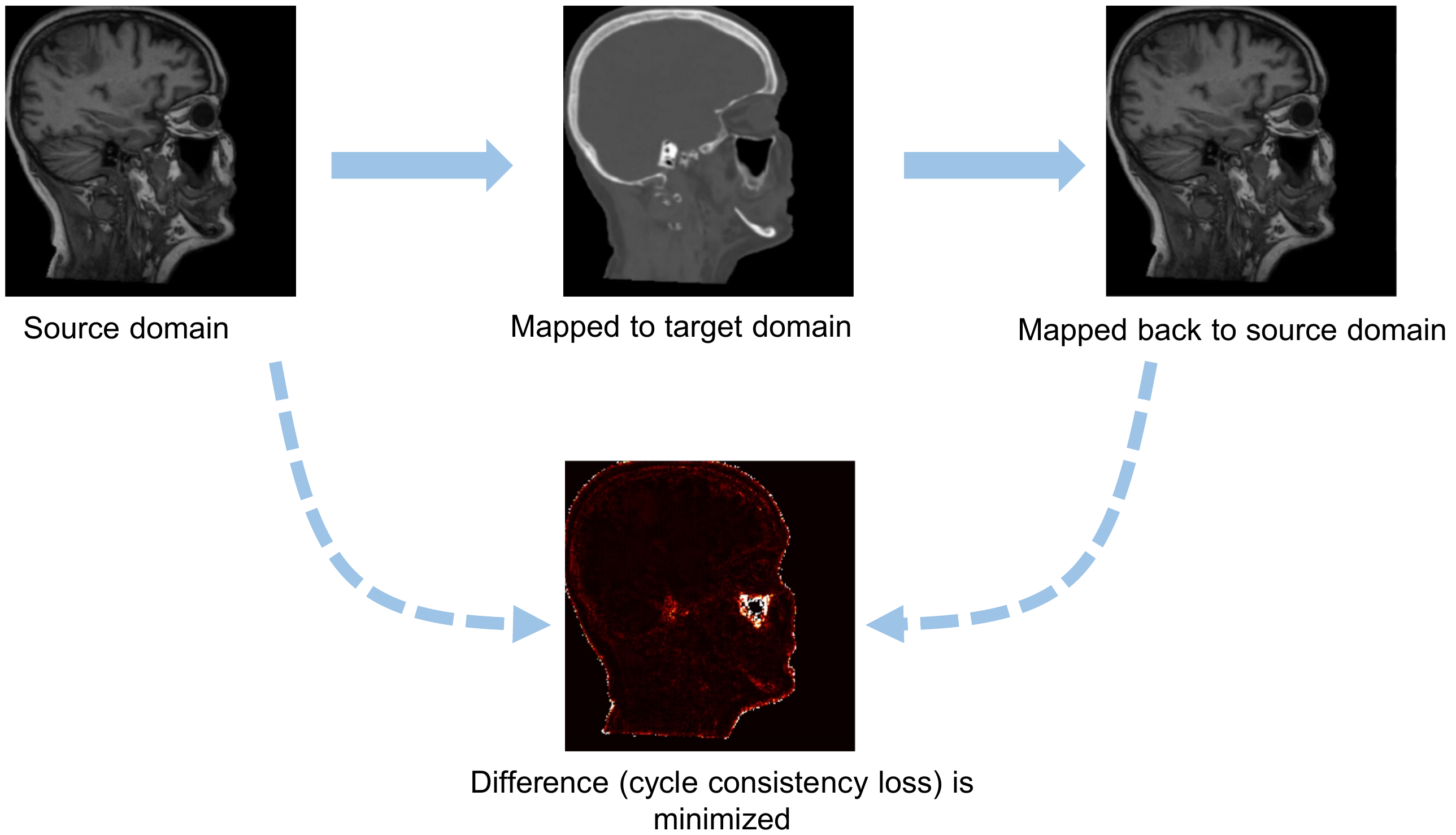}
 \caption{Image-to-image translation via cycle consistency loss (CycleGAN)~\cite{cycleGAN,cycleGAN_medical}. The source domain MR image is mapped to the target domain CT image, and then mapped back to the source domain. The difference between the input MR image and the reconstructed MR image is minimized.}
\label{fig:cycleGAN}
\end{figure}

Image-level alignment is also used for domain adaptation based on deep generative models, such as Generative Adversarial Network (GAN)~\cite{GAN}. 
The original GAN takes random noise as input. 
As a significant improvement, Zhu~\etal\cite{cycleGAN} propose a cycle-consistent GAN (CycleGAN) model which can translate one image domain into another without the demand for paired training samples. 
In medical image analysis, CycleGAN has been used for image synthesis (see Fig.~\ref{fig:cycleGAN}), where an MR image is firstly mapped to the target domain (CT image) and then mapped back to the input domain. 
A cycle consistency loss is used to measure the difference between the input image and the reconstructed image. Through minimizing this loss, the CycleGAN can realize image-to-image translation without paired training samples. 

Wollmann~\etal\cite{CycleGAN_Dense} propose a CycleGAN-based DA method for breast cancer classification. They first use CycleGAN to transform  whole-slide images (WSIs) of lymph nodes from a source domain (\ie, a medical center) to the target domain (\ie, another medical center). 
Then a densely connected deep neural network (DenseNet) is used for breast cancer classification. 
Image patches from region-of-interest (ROIs) are used as the input to facilitate two-level classification.
Experiments on the CAMELYON17 challenge dataset\footnote{https:
//camelyon17.grand-challenge.org/} demonstrate the effectiveness.
Manakov~\etal\cite{DART19_noise} propose to leverage unsupervised DA to tackle retinal optical coherence tomography (OCT) image denoising problem. 
They treat image noises as domain shift between high and low noise domains.
A CycleGAN-based DA mode is designed to learn a mapping between the source (\ie, high noise) and the target domain (\ie, low noise) on unpaired OCT images, thus achieving the goal of image denoising. 
Experiments on numerous in-house OCT images demonstrate the effectiveness of this method.
Gholami~\etal\cite{tumor_GAN} propose to use CycleGAN to generate more training data for brain tumor segmentation. They first generate synthetic tumor-bearing MR images using their in-house simulation model, and then transform them to real MRIs using CycleGAN to augment training samples. 
They conduct experiment on the BraST~\cite{BRATS} dataset and achieve good segmentation results for brain tumors.

Zhang~\etal\cite{Noise_GAN} propose a Noise Adaptation Generative Adversarial Network (NAGAN) for eye blood vessel segmentation, by formulating DA as a noise style transfer task. 
They use an image-to-image translation generator to map the target image to the source domain. 
Besides a normal discriminator which enforces the content similarity between generated images and real images, they also design a style discriminator which enforces the generated images to have the same noise patterns as those from the target domain. 
Experiments on the SINA~\cite{SINA} and a local dataset demonstrate the effectiveness.
Mahmood~\etal\cite{Reverse_GAN} propose a Reverse Domain Adaptation method based on GAN for endoscopy image analysis. 
Different from traditional GAN, they reverse the flow and transform real images to synthetic images, based on the insight that subjects with optical endoscopy images have patient-specific details and these details do not generalize across patients. 
To remove these details from real images while maintaining useful diagnostic details, they transform real images in the target domain to synthetic ones, thus bridging the gap between the target and source domains. 

\subsubsection{Image+Feature Alignment}
Chen~\etal\cite{SIFA,SIFA_2} combine image alignment and feature alignment for cross-modality cardiac image segmentation. 
They firstly use CycleGAN to transform labeled source images into target-like images. 
Then, the synthesized target images and real target images are fed into a two-stream CNN with a domain discriminator which can further reduce the domain gap via adversarial learning.
Experiments on the MM-WHS dataset~\cite{MM-WHS} demonstrate its effectiveness.
Yan~\etal\cite{Unet-GAN} propose a similar unsupervised framework for cross-vendor cardiac cine MRI segmentation. 
They first train a U-Net using labeled source samples, and then train a CycleGAN to facilitate image-level adaptation. 
In addition, they design a feature-level alignment mechanism by computing the Mean Square Error (MSE) of features between the original and translated images. 
Three major vendors including Philips, Siemens, and GE, are treated as three domains. For example, the Philips data are treated as the source domain, while the Siemens and GE data are defined as two target domains. 
The short-axis steady-state free precession (SSFP) cine MR
images of 144 subjects acquired by those three vendors are used for performance evaluation. 

\subsubsection{Disentangled Representation}
Yang~\etal\cite{Disentangle} propose a cross-modality (between CT and MRI) DA method via disentangled representation for liver segmentation. 
Through disentangle presentation learning, images from each domain are embedded into two spaces, \ie, a shared domain-invariant content space and a domain-specific style space. 
Then, domain adaptation is performed in the domain-invariant space.
Experiments on the LiTS benchmark dataset~\cite{LiTS} and a local dataset demonstrate the effectiveness of this method.

\subsubsection{Ensemble Learning}
Following~\cite{Self_Ensemble}, Perone~\etal\cite{Self_Ensemble_1} propose a self-ensemble based DA method for medical image segmentation. 
A baseline CNN, called student network, takes labeled samples from the source domain, and makes its predictions after training with a task (segmentation) loss. 
Another network, called teacher network, gives its predictions using only unlabeled samples in the target domain. 
The teacher network is updated by the exponential moving average of the weights of the student network (a temporal ensemble strategy). 
During training, the domain shift is minimized by a consistency loss by comparing predictions from both the student and teacher networks. This method is evaluated on the SCGM challenge dataset~\cite{SCGM}, which is a multi-center multi-vendor dataset of spinal cord anatomical MR images from healthy subjects.  
%
Shanis~\etal\cite{Self_Ensemble_2} employ a similar DA architecture for brain tumor segmentation. Besides a consistency loss that measures the difference between predictions of the teacher and student networks, they also utilize an adversarial loss to further improve adaptation performance.
Effectiveness is validated with experiments on the BraTS dataset~\cite{BRATS}.

\subsubsection{Soft Labels}
Bermúdez-Chacón~\etal\cite{Visual_Corr} propose an unsupervised DA method to reduce domain shift in electron microscopy images.
Based on the observation that some ROIs in medical images still resemble each other across domains, the authors propose to use Normalized Cross Correlation (NCC) which computes the similarity of two structures from two different images (visual correspondence) to construct a consensus heatmap. 
The high-scoring regions of the heatmap are used as soft labels for the target domain, which are then used to train the target classifier for segmentation. 

\subsubsection{Feature Learning}
Ahn~\etal\cite{CAE2,CAE} introduce a convolutional auto-encoder to learn features for domain adaptation. 
Other than using labeled target data to fine-tune a pre-trained AlexNet, the authors propose to use a zero-bias convolutional auto-encoder as an adapter to transform the feature maps from the last convolution layer of AlexNet to relevant medical image features. 
The learned features can further be used for classification tasks. 
Experimental results on the ImageCLEF dataset\footnote{https://www.imageclef.org/} and ISIC dataset~\cite{ISIC} validate the effectiveness of this method.

\subsection{Multi-Target Deep DA}
Orbes-Arteaga~\etal\cite{Pair_Consistency} propose an unsupervised DA framework with two target domains for brain lesion segmentation. 
Their network is first trained with labeled source data. During the adaptation phase, labeled source data and the paired unlabeled data from the two target domains are fed into the DA model. 
An adversarial loss is imposed to reduce domain differences. To minimize the output probability distribution on the target data, a consistency loss is imposed on the paired target data.
Experiments are performed on the MICCAI 2017 WMH Challenge~\cite{WMH_Challenge} and two independent datasets. 
Karani~\etal\cite{Multi_DA_BN} propose a life-long multi-domain adaptation framework for brain structure segmentation.
In their method, batch normalization plays the key role in adaptation, as suggested in~\cite{BN_1}.
They first train a network on $d$ source domains with shared convolutional filters but different batch normalization (BN) parameters, and each domain corresponds to a specific set of BN parameters. 
By fine-tuning the BN parameters with a few labeled data in each new target domain, the model can then be adapted to those multi-target domains.
This method is evaluated on 4 publicly available datasets, including HCP~\cite{HCP}, ADNI~\cite{ADNI}, ABIDE~\cite{ABIDE}, and IXI\footnote{brain-development.org/ixi-dataset/}. 

\begin{table*}[t]
\setlength{\belowcaptionskip}{-2pt}
\setlength{\abovecaptionskip}{-2pt}
\setlength{\abovedisplayskip}{-2pt}
\setlength{\belowdisplayskip}{-2pt}
\caption{Overview of benchmark datasets for domain adaptation research on different medical image analysis tasks.}
\centering
\setlength{\tabcolsep}{1mm}{
\begin{tabular} {lllll} 
\toprule
Reference  &Task   &Dataset   &Image Type   &Method\\
\midrule
\textbf{Brain}          &~         &~         &~           &~\\
Wachinger~\etal\cite{Reweighting_Wachinger}  &AD classification        &ADNI, AIBL,  CADDementia      &MRI           &Instance weighting\\

Cheng~\etal\cite{DTSVM_2}  &MCI conversion prediction        &ADNI       
&MRI, PET           &Adaptive SVM\\

Cheng~\etal\cite{Multiple_Domain_MCI}  &AD classification         &ADNI       
&MRI          &Ensemble, Adaptive SVM\\

Li~\etal\cite{AD_SVD}  &AD classification       &ADNI       &fMRI          &Subspace alignment\\

Hofer~\etal\cite{Simple_DA} &AD classification       &ADNI, IXI       &MRI          &Instance alignment\\

Wang~\etal\cite{ASD_MDA_1}  &ASD classification       &ABIDE       &fMRI          &low-rank regularization\\

Wang~\etal\cite{ASD_MDA_2}  &ASD classification       &ABIDE       &fMRI          &low-rank regularization\\

Goetz~\etal\cite{re-weighting} &Tumor segmentation        &BraTS      
&MRI          &Instance weighting\\

Kaur~\etal\cite{Different_Disease_TL} &Tumor segmentation        &BraTS      
&MRI          &CNN fine-tuning\\

Gholami~\etal\cite{tumor_GAN} &Tumor segmentation        &BraTS      
&MRI          &GAN\\

Shanis~\etal\cite{Self_Ensemble_2}  &Tumor segmentation        &BraTS      
&MRI          &CNN, feature alignment\\

Swati~\etal\cite{VGG_MRI}  &Tumor classification       &CE-MRI     
&MRI          &CNN fine-tuning\\

Karani~\etal\cite{Multi_DA_BN}  &Structure segmentation        &HCP, ADNI, ABIDE, IXI      &MRI          &CNN, feature alignment\\

Khan~\etal\cite{Khan}  &AD classification        &ADNI    
&MRI          &CNN fine-tuning\\

Hosseini-Asl~\etal\cite{Adaptive_CNN}  &AD classification        &ADNI, CADDementia     &MRI          &CNN fine-tuning\\

Zhang~\etal\cite{UCAN}    &AD classification        &ADNI  &MRI          &Adversarial learning\\

Guan~\etal\cite{AD2A}    &AD classification        &ADNI, AIBL  &MRI          &CNN, feature alignment\\

Mahapatra~\etal\cite{register_GAN}   &Registration       &ADNI  &MRI          &CNN, feature alignment\\

Valverde~\etal\cite{FC_DA}  &Lesion segmentation        &ISBI2015   &MRI          &CNN fine-tuning\\

Orbes-Arteaga~\etal\cite{Pair_Consistency}  &Lesion segmentation        &MICCAI WMH Challenge  &MRI          &CNN, feature alignment\\

Gao~\etal\cite{fMRI_CMD}  &Brain activity classification    &HCP  &fMRI          &CNN, feature alignment\\

\hline
\textbf{Lung}          	&~     &~         &~           &~\\
Cheplygina~\etal\cite{Lung_Weighting} &COPD classification        &DLCST, COPDGene        &CT           &Instance weighting\\

Yang~\etal\cite{ROI_DANN} &Lung texture classification        &SPIROMICS  &CT          &CNN, feature alignment\\

Mahapatra~\etal\cite{register_GAN}   &Registration        &NIH ChestXray14  &X-ray          &CNN, feature alignment\\
\hline
\textbf{Heart}          	&~     &~         &~           &~\\
Madani~\etal\cite{Semi_GAN}   &Cardiac abnormalities identity         &NIH PLCO, NIH Chest       &X-ray           &GAN\\

Dou~\etal\cite{Early_DA} &Cardiac segmentation      &MM-WHS       &CT, MRI           &CNN, feature alignment\\

Chen~\etal\cite{SIFA}  &Cardiac segmentation      &MM-WHS       &CT, MRI           &GAN, feature alignment\\
\hline
\textbf{Breast}       	&~          &~       &~         &~\\
Kumar~\etal\cite{ResNet_TCA}  &Cancer classification        &CBIS-DDSM, InBreast, MIAS        &X-ray           &CNN feature, shallow\\

Shen~\etal\cite{mammogram_Shen}  &Mass detection        &CBIS-DDSM, InBreast    &X-ray           &CNN, feature alignment\\

Wollmann~\etal\cite{CycleGAN_Dense} &Cancer classification      &CAMELYON17    &Histological image          &GAN\\

\hline
\textbf{Skin}         	&~        &~          &~           &~\\
Gu~\etal\cite{Two_Step_TL}  &Skin disease classification        &MoleMap, HAM10000        &dermoscopic, camera images           &GAN\\

Ahn~\etal\cite{CAE}  &Skin disease classification        &ISIC  &dermoscopic images           &CNN, feature alignment\\
\hline
\textbf{Eye}         		&~        &~         &~           &~\\
Javanmardi~\etal\cite{DANN_Eye}   &Retina segmentation        &DRIVE, STARE        &Color retinal images             &CNN, feature alignment\\

Zhang~\etal\cite{Noise_GAN}   &Vasculture segmentation        &SINA, AROD        &OCT             &GAN\\

\hline
\textbf{Abdomen}   &~        &~         &~          &~\\
Zhu~\etal\cite{BOWDA}  &Prostate segmentation        &PROMISE12   &MRI           &CNN fine-tuning\\

Yang~\etal\cite{Disentangle}  &Liver segmentation       &LiTS   &CT, MRI  &GAN\\
\hline
\textbf{Histology and Microscopy}         &~        &~         &~         &~\\
Huang~\etal\cite{Huang}  &Epithelium-stroma classification        &NKI, VGH and IHC        &Histological image         &CNN fine-tuning\\
\hline
\textbf{Others}   &~        &~         &~         &~\\
Bateson~\etal\cite{Constrained_DA}   &Disc segmentation       &MICCAI2018 IVDM3Seg        &MRI        &CNN, feature alignment\\

Perone~\etal\cite{Self_Ensemble_1}  &Spinal cord gray matter segmentation       &SCGM Challenge        &MRI        &CNN, feature alignment\\

Shanis~\etal\cite{Self_Ensemble_2}  &Spinal cord gray matter segmentation       &SCGM Challenge        &MRI        &CNN, feature alignment\\
\bottomrule
\end{tabular}
}
\label{tab:dataset}
\end{table*}

\section{Benchmark Datasets}  \label{Datasets}
We now introduce benchmark datasets for domain adaptation in medical image analysis (see Table~\ref{tab:dataset}) in terms of different research objects/organs.

\subsection{Brain Images}
\subsubsection{ADNI}
The Alzheimer's Disease Neuroimaging Initiative (ADNI) is the most influential project for the research of Alzheimer's disease (AD)~\cite{ADNI,ADNI3}. 
It involves four datasets, \ie, ADNI-1, ADNI-2, ADNI-GO and ADNI-3, where MRI, PET and fMRI are the most popular modalities in DA research.

\subsubsection{AIBL}
The Australian Imaging, Biomarkers and Lifestyle (AIBL)~\cite{AIBL} dataset is a benchmark for AD research, including MRI and PET data. All the data are collected from two medical centers in Australia.

\subsubsection{CADDementia}
The Computer-Aided Diagnosis of Dementia (CADDementia) dataset~\cite{CADDementia} consists of clinical-representative T1-weighted MRIs of patients from multiple centers with Alzheimer's disease (AD), mild cognitive impairment (MCI) and healthy controls.  

\subsubsection{IXI}
The Information eXtraction from Images (IXI) datasetIXI\footnote{brain-development.org/ixi-dataset/} contains 600 MR images from normal, healthy subjects, collected from three different hospitals in London with different scanners.  

\subsubsection{ABIDE}
Autism Brain Imaging Data Exchange (ABIDE) is a benchmark for the research of Autism~\cite{ABIDE}. It has aggregated numerous functional and structural brain imaging data from more than 24 imaging centers around the world.

\subsubsection{ISBI2015}
The ISBI2015 MS lesion challenge~\cite{ISBI2015} is a public dataset for lesion segmentation. It consists of 5 training and 14 testing subjects, and each subject has 4 or 5 different images at different time-points.

\subsubsection{BraTS}
The multi-modal brain tumor image segmentation benchmark (BraTS) is a popular dataset for brain tumor segmentation tasks~\cite{BRATS}. The MR images are collected from four different centers with different scanners.

\subsubsection{MICCAI WMH Challenge}
The MICCAI White Matter Hyperintensities (WMH) Challenge dataset~\cite{WMH_Challenge} is a collection of brain MR images, consisting of 60 training and 110 test images. All the imaging data are acquired from five different scanners in three different institutes.

\subsubsection{CE-MRI dataset}
The T1-weighted contrast-enhanced magnetic resonance images (CE-MRI) benchmark dataset\footnote{https://figshare.com/articles/brain\_tumor\_dataset/1512427} is a brain tumor dataset which consists of 3,064 T1 contrast-enhanced MRIs with three categories of brain tumor. 

\subsubsection{HCP}
The Human Connectome Project (HCP)~\cite{HCP}. is a large database of brain research data provided by the National Institutes of Health (NIH). The dataset consists of 1,200 subjects which are scanned in four modalities, \ie, structural MRI, resting-state fMRI, task fMRI and diffusion MRI, on a 3T scanner at Washington University (WashU). A subset with 200 subjects is scanned on a 7T scanner at the University of Minnesota (UMinn). 

\subsection{Lung Images}
\subsubsection{NIH ChestXray14}
The NIH ChestXray14~\cite{ChestXray14,register_GAN} is a large-scale public dataset, consisting of 112,120 frontal-view X-rays. There are totally 14 categories of disease (multi-labels for each image). 

\subsubsection{DLCST}
The Danish Lung Cancer Screening Trial (DLCST)~\cite{DLCST} contains CT images for lung cancer research.

\subsubsection{COPDGene}
The COPDGene dataset~\cite{COPDGene} contains chest CT images, and all images are acquired from 21 clinical study centers in the United States.
\subsection{Heart Images}
\subsubsection{MM-WHS}
The MICCAI 2017 Multi-Modality Whole Heart Segmentation challenge (MM-WHS) dataset~\cite{MM-WHS} provides 20 MR and 20 CT cardiac images with pixel-wise segmentation annotations. The images are unpaired and come from different sites, which enables it suitable for cross-modality adaptation research.

\subsubsection{NIH PLCO}
The National Institute of Health (NIH) Prostate, Lung, Colorectal, and Ovarian (PLCO) cancer dataset~\cite{PLCO} consists of about 196,000 X-ray images acquired from 10 imaging centers throughout the United States.

\subsubsection{NIH Chest}
The NIH Chest X-Ray dataset~\cite{NIH_Chest} is a collection of chest X-ray images. It consists of 8,121 images acquired from 2 large hospital systems within the Indiana Network by Indiana University.

\subsection{Eye Images}
\subsubsection{DRIVE}
The DRIVE dataset~\cite{DRIVE} is a popular eye blood vessel segmentation dataset. It contains 40 color images with corresponding pixel-level ground truth. All the images are divided into a training and test set, each containing 20 images.

\subsubsection{STARE}
The STructured Analysis of the Retina (STARE) dataset\footnote{https://cecas.clemson.edu/~ahoover/stare/} is a popular project for retinal imaging. It consists of 400 eye images, with partially labeled data for blood vessel segmentation. 

\subsubsection{SINA}
The SINA~\cite{SINA} is a publicly optical coherence tomography (OCT) dataset. 
It has 220 B-scans from 20 volumes of eyes, collected using OCT imaging systems.

\subsection{Breast Images}
\subsubsection{CBIS-DDSM}
The Curated Breast Imaging Subset of DDSM (CBISD-DSM)~\cite{DDSM} is a publicly available dataset with 3,103 scanned mammogram images. All the images are labeled as Benign or Malignant with verified pathology information. Most of the mammogram images have cranial cardo (CC) and media later oblique (MLO) views.

\subsubsection{InBreast}
The InBreast~\cite{InBreast} is a public full-field digital mammography dataset with 10 mammogram images (labeled as benign or malignant) from 115 patients. It includes BI-RAIDS readings for images, but lacks biopsy confirmation. 

\subsubsection{CAMELYON} 
The CAMELYON challenge dataset\footnote{https:
//camelyon17.grand-challenge.org} contains histological images from 100 patients collected by five different medical centers in the Netherlands.

\subsubsection{MIAS} 
The MIAS dataset~\cite{MIAS} has 322 mammogram images with MLO view. All the images are scanned copies of films with the size of $1024\times1024$.

\subsection{Skin Images}
\subsubsection{MoleMap}
The MoleMap dataset\footnote{http://molemap.co.nz} consists of 102,451 skin images. There are totally 25 skin conditions in these images from 3 cancerous categories and 22 benign categories.

\subsubsection{HAM10000}
The Human Against Machine with 10,000 training images (HAM10000) dataset\footnote{https://www.kaggle.com/kmader/skin-cancer-mnist-ham10000} is a large collection of multi-source dermatoscopic images of skin lesions.
It consists of 10,015 dermatoscopic images with 7 skin categories, including 2 cancerous categories and 5 benign categories.

\subsubsection{ISIC}
The International Skin Imaging Collaboration (ISIC) challenge dataset~\cite{ISIC} is a public dermoscopic skin image dataset. It houses 20,000 images in 3 skin condition categories that are acquired from several centers.

\subsection{Abdomen Images}
\subsubsection{PROMISE12}
The MICCAI 2012 Prostate MR Image Segmentation challenge dataset (PROMISE12)\footnote{https://promise12.grand-challenge.org/} is a publicly available dataset for evaluating prostate segmentation  algorithms from MR images. Overall, it consists of 50 T2-weighted MRIs of the prostate and the corresponding segmentation ground truth acquired in different hospitals.

\subsubsection{BWH}
The Brigham and Women’s Hospital (BWH) Multiparametric MR (mpMR) prostate image dataset~\cite{BWH} is a collection of prostate MRIs acquired from 15 subjects.

\subsubsection{LiTS}
The Liver Tumor Segmentation Benchmark (LiTS) is a publicly available dataset for liver segmentation. It consists of 201 CT scans of patients, acquired by several hospitals and research institutions. Annotations are manually reviewed independently by three radiologists.

\subsection{Histology and Microscopy Images}
\subsubsection{NKI, VGH}
The Netherland Cancer Institute (NKI) and Vancouver General Hospital (VGH) datasets are collected independently by the Netherland Cancer Institute (248 patients, 778 images) and Vancouver General Hospital (328 patients, 666 images)~\cite{NKI_VGH}.
Both NKI and VGH consist of histological images from breast cancer tissue microarrays, and each image has the size of $1128 \times 720$. 

\subsubsection{IHC}
The Immunohistochemistry Dataset (IHC)~\cite{IHC} contains 1,377 rectangular tissue samples from 643 patients with colorectal cancer at Helsinki University, and each sample is labeled as epithelium or stroma.

\subsection{Others}
\subsubsection{MICCAI2018 IVDM3Seg}
This is a public dataset for disc localization and segmentation\footnote{https://ivdm3seg.weebly.com}, containing 16 multi-modality MRIs of lower spine with manual segmentations.

\subsubsection{SCGM}
The Spinal Cord Gray Matter Challenge (SCGM) dataset~\cite{SCGM} is a publicly-available MRI data collection. It contains 80 healthy subjects from 4 medical centers.


\section{Discussion}  \label{Discussions}

\subsection{Challenges of Data Adaptation for Medical Image Analysis} 

\subsubsection{3D/4D Volumetric Representation}
Medical images are generally high-dimensional and stored in 3D or 4D formats (\eg, with the temporal dimension). 
Especially for time-series data, \eg, function MRI, there are a series of 3D volumes for each subject, and each 3D volume is composed of hundreds of 2D image slices. 
These slices usually contain rich structural context information that is significant for representing medical images. 
It is challenging to design advanced domain adaptation models to effectively capture the 3D or 4D structural information convened in medical images.

\subsubsection{Limited Training Data}
The existing medical image datasets usually contain a limited number of samples, ranging from a few hundred~\cite{AIBL} to several hundred thousand~\cite{ChestXray14}. 
Also, labeled medical images are generally even fewer, since labeling medical images is a time-consuming and expensive task which demands for the participation of medical experts. 
Even though one can adapt/transfer models pre-trained on large-scale ImageNet, the existing off-the-shelf deep models may not be well adapted to medical images, since they are designed for 2D image analysis and have a huge number of parameters at the higher layers for classification of a large number of categories (\eg, $>1,000$). 
Raghu~\etal\cite{Transfusion} conduct an empirical study on two large-scale medical imaging datasets, and find that transferring standard CNN models pre-trained on ImageNet to medical imaging analysis does not benefit performance a lot. 
The problem of limited data has posed a great challenge to effective training of domain adaptation models, especially for deep learning-based ones.

\if false
\subsubsection{Definition of Task-Specific ROIs}
In each medical image, there may contain a lot of redundant or noisy regions, while task-specific ROIs can help filter out these redundant/noisy regions. 
For instance, brain structural changes caused by AD usually locate in a limited number of brain regions in structural MRI scans (rather than the whole-brain MRI), such as the hippocampus~\cite{Hippocampus} and ventricles~\cite{Ventricles}. 
This implies that a large number of brain regions may not discriminative for MRI-based AD identification.
However, effectively defining these task-specific ROIs in medical images for domain adaptation has always been an open issue. {\color{blue} This item is not mentioned in Future Trends}
\fi

\subsubsection{Inter-Modality Heterogeneity}
Various digital imaging techniques have been developed to generate heterogeneous visual representations of each subject, such as CT, structural MRI, function MRI, and positron emission tomography (PET). 
While these multi-modality data provide complementary information, the inter-modality heterogeneity brings many challenges for domain adaptation~\cite{SIFA}. 
For instance, a pair of structural MRI and PET scans from the same subject contains large inter-modality discrepancy~\cite{Yang}. 
The large inter-modality difference brings many difficulties actually for efficient knowledge transfer between different domains.

\if false
\section{Discussions}  \label{Discussions}
\subsection{Challenges of Data Adaptation for Medical Image Analysis} 
Domain adaptation has made great progress in natural image analysis, thanks to the large-scale natural image datasets. 
Leveraging models pre-trained on natural images is a popular way to tackle the domain shift problem in medical imaging analysis. 
However, there are fundamental differences between natural images and medical images in terms of data format, dataset size, and task specification. 
Due to the specific property of medical images, there are still many challenges to further advance domain adaptation techniques for medical image analysis.

\subsubsection{3D/4D Volumetric Representation}
Medical images are often in 3D or 4D format (with the temporal dimension) and each 3D/4D volume is composed of hundreds of 2D image slices. 
Also, they usually contain significantly fewer classes than natural images. 
Thus, the existing off-the-shelf CNN architectures pre-trained on ImageNet may not be well adapted to medical images, since they are designed for 2D image analysis and have a huge number of parameters at the higher layers for classification of a large number of categories (\eg, >1,000). 
For example, Raghu~\etal\cite{Transfusion} conduct an empirical study on two large-scale medical imaging datasets, and find that transferring standard CNN models pre-trained on ImageNet to medical imaging analysis does not benefit performance a lot. 
It is challenging to design advanced domain adaptation models to not only borrow information learned on 2D natural images but also effectively capture the 3D- or 4D structural information convened in medical images. {\color{cyan}Strange logic here.}

\subsubsection{Limited Labeled Training Data}
Additionally, compared with large-scale labeled natural image datasets, \eg, ImageNet, medical image datasets have much fewer images, ranging from a few hundred~\cite{AIBL} to several hundred thousand~\cite{ChestXray14}. 

In addition, since labeling medical images is a time-consuming and expensive task which demands for the participation of medical experts, labeled medical image are even fewer. This has posed a great challenge to adaptation model training, especially the deep learning-based ones.

\subsubsection{Task-specific Regions of Interest}
Finally, unlike natural images, the most informative parts are often within a small region of a medical image. 
For example, in MR images for AD diagnosis, the most discriminative regions mainly locate in the hippocampus~\cite{Hippocampus} and ventricles~\cite{Ventricles}. 
This is quite different from natural images where there is often a global object,~\eg, face recognition. 
How to effectively explore these important regions during the domain adaptation and avoid noises caused by unrelated areas process remains an open problem.
\fi

\subsection{Future Research Trends}
\subsubsection{Task-Specific 3D/4D Models for Domain Adaptation}
Compared with 2D CNNs pre-trained on ImageNet, 3D models are usually more powerful to explore medical image features and yield better learning performance~\cite{3D_medical}. 
For more high-dimensional medical images such as fMRI that involves temporal information, a 4D CNN has been introduced in the literature~\cite{4D_CNN_fMRI}. 
Besides, medical images usually contain redundant or noisy regions, while task-specific ROIs can help filter out these redundant/noisy regions. 
For instance, brain structural changes caused by AD usually locate in a limited number of ROIs in structural MRIs, such as hippocampus~\cite{Hippocampus} and ventricles~\cite{Ventricles}. 
However, effectively defining these task-specific ROIs in medical images has always been an open issue. 
Currently, there are very few works on developing 3D/4D DA models with task-specific ROI definitions for medical image analysis, and we believe this is a promising future direction. 

\subsubsection{Unsupervised Domain Adaptation}
The lack of labeled data is one of the most significant challenges for medical image research. 
To tackle this problem, many recent studies tend to avoid using labeled target data for model fine-tuning, by using various unsupervised domain adaptation methods~\cite{DA_3,DA_4} for medical image analysis. 
Also, completely avoiding any target data (even those unlabeled ones) for model training is an interesting research topic. 
Thus, research on domain generalization~\cite{DA_G,domain-general-1,domain-general-2,domain-general-3} and zero-shot learning~\cite{zero-shot-1,zero-shot-2,zero-shot-3} will be welcome in the future. 

\if false
\subsubsection{Multi-Source Multi-Target Domain Adaptation}
Existing DA methods usually focus on single-source domain adaptation, \ie, training a model on one source domain, but there may be multiple source domains (\eg, multiple imaging centers) in real-world applications. 
Multi-source domain adaptation~\cite{MDA_1,MDA_2,Multi_DA1,Multi_DA2}, aiming to utilize training data from multi-source domains to improve models' transferability on the target domain, is of great clinical significance. 
It is also practical to transfer a model to multiple target domains, \ie, multi-target DA. 
Currently, very limited works have been done on multi-source/multi-target DA in medical image analysis, so there is still a lot of room for future research.
\fi

\subsubsection{Multi-Modality Domain Adaptation}
To make use of complementary but heterogeneous multi-modality neuroimaging data, it is desired to develop domain adaptation models that are trained on one modality and can be well generalized to another modality. 
For more challenging problems where each domain contains multiple modalities (\eg, MRI and CT), it is meaningful to consider both inter-modality heterogeneity and inter-domain difference when designing domain adaptation models. 
Several techniques including CycleGAN~\cite{SIFA,cycleGAN_medical} and disentangle learning~\cite{Disentangle,Yang} have been introduced into this emerging area, while further exploration of multi-modality DA is required for medical image analysis. 

\subsubsection{Multi-Source/Multi-Target Domain Adaptation}
Existing DA methods usually focus on single-source domain adaptation, \ie, training a model on one source domain, but there may be multiple source domains (\eg, multiple imaging centers) in real-world applications. 
Multi-source domain adaptation~\cite{MDA_1,MDA_2,Multi_DA1,Multi_DA2}, aiming to utilize training data from multi-source domains to improve models' transferability on the target domain, is of great clinical significance. 
It is also practical to transfer a model to multiple target domains, \ie, multi-target DA. 
Currently, very limited works have been done on multi-source/multi-target DA in medical image analysis, so there is still a lot of room for future research.

\section{Conclusion} \label{Conclusion}
In this paper, we provide a survey of recent advances in domain adaptation for medical image analysis. 
We categorize existing methods into shallow and deep models, and each category is further divided into supervised, semi-supervised and unsupervised methods. 
We also introduce the existing benchmark medical image datasets used for domain adaptation. 
We summarize potential challenges and discuss future research directions. 
We hope this survey can offer the readers a better understanding of the current DA-related studies and future directions in this promising research area.


\ifCLASSOPTIONcaptionsoff
  \newpage
\fi

\footnotesize
\bibliographystyle{IEEEtran}
\bibliography{mybib}

\end{document}

%% file: defs.tex
\def\0{{\bf 0}}
\def\1{{\bf 1}}

\def\etal{{\em et al.}}
\def\eg{{\em e.g.}}
\def\ie{{\em i.e.}}
\def\etc{{\em etc}}

\def\etal{{\em et al.\/}\,}
